\documentclass[10pt, conference, compsocconf]{IEEEtran}
\usepackage{array}
\usepackage{mathtools}
\usepackage{subcaption}
\usepackage{amssymb}
\usepackage{amsmath}
\usepackage{diagbox}
\usepackage{fourier} 
\usepackage{flushend}
\usepackage[linesnumbered,ruled]{algorithm2e}
\usepackage{array}
\DeclareMathAlphabet{\mathcal}{OMS}{cmsy}{m}{n}
\usepackage{makecell}
\usepackage{hyperref}
\usepackage{stackengine}
\usepackage{multirow}
\usepackage[symbol]{footmisc}
\usepackage{sidecap}
\usepackage{tablefootnote}
\usepackage[flushleft]{threeparttable}
\usepackage{booktabs,caption}
\usepackage{xcolor, soul}
\sethlcolor{yellow}

\DeclareMathOperator*{\argmax}{argmax}

\begin{document}
\title{Next-Best-View Selection for Robot Eye-in-Hand Calibration}


\author{\IEEEauthorblockN{Jun Yang, Jason Rebello, Steven L. Waslander}
\IEEEauthorblockA{Institute for Aerospace Studies and Robotics Institute\\
University of Toronto\\
Toronto, Canada\\
\{jun.yang, jason.rebello, steven.waslander\}@robotics.utias.utoronto.ca}
}

\maketitle

\begin{abstract}
Robotic eye-in-hand calibration is the task of determining the rigid 6-DoF pose of the camera with respect to the robot end-effector frame. In this paper, we formulate this task as a non-linear optimization problem and introduce an active vision approach to strategically select the robot pose for maximizing calibration accuracy. Specifically, given an initial collection of measurement sets, our system first computes the calibration parameters and estimates the parameter uncertainties. We then predict the next robot pose from which to collect the next measurement that brings about the maximum information gain (uncertainty reduction) in the calibration parameters. We test our approach on a simulated dataset and validate the results on a real 6-axis robot manipulator. The results demonstrate that our approach can achieve accurate calibrations using many fewer viewpoints than other commonly used baseline calibration methods.

\end{abstract}


\IEEEpeerreviewmaketitle

\section{Introduction}
With the rapid advance in 3D computer vision techniques, perception systems have become an essential component in many industrial applications, such as robotic bin-picking~\cite{abbeloos2016point,chen2018random} and on-machine inspection~\cite{phan2019optimal}. Historically, for these applications, the sensor is installed in a static position. For example, in robotic bin-picking, the camera is mounted above the bin. However, the camera often fails to acquire a complete representation of the task space from a single viewpoint due to occlusions and limited sensor resolution. To overcome these limitations, we can attach the camera to the end-effector of a robotic manipulator, which is also known as the eye-in-hand system.

The proper functioning of an eye-in-hand system relies on accurate calibration. The goal is to estimate a rigid 6-DoF pose of the camera with respect to the robot end-effector frame. As introduced in~\cite{tsai1989new,shiu1987calibration}, this problem can be solved with a hand-eye calibration formulation $\mathbf{AX=XB}$, where $\mathbf{A}$ and $\mathbf{B}$ represent the relative robotic arm and camera motions between two different time instants, respectively, and $\mathbf{X}$ is the unknown rigid transformation from the robot end-effector frame to the camera coordinate frame. To use this formulation, we need to convert absolute poses into relative ones and solve the $\mathbf{AX=XB}$ problem using closed-form approaches~\cite{tsai1989new, shiu1987calibration} or optimization-based solutions~\cite{horaud1995hand}. Alternatively, the eye-in-hand calibration can be solved with the robot-world-hand–eye formulation $\mathbf{AX=YB}$~\cite{zhuang1994simultaneous}. In this formulation, $\mathbf{Y}$ represents the transformation from the robot end-effector to the camera frame, and $\mathbf{X}$ is the transformation from the robot base to the world frame. The transformations $\mathbf{A}$ and $\mathbf{B}$ represent the absolute transformation from the robot base to the robot end-effector, as well as the transformation from the camera to the world coordinate. The goal of robot-world-hand–eye calibration is to determine both $\mathbf{X}$ and $\mathbf{Y}$ from either closed-form solutions~\cite{zhuang1994simultaneous,dornaika1998simultaneous,aiguo2010simultaneous} or optimization-based approaches~\cite{remy1997hand,tabb2017solving}.

\begin{figure}[t]
\centering
  \includegraphics[width=0.8\linewidth]{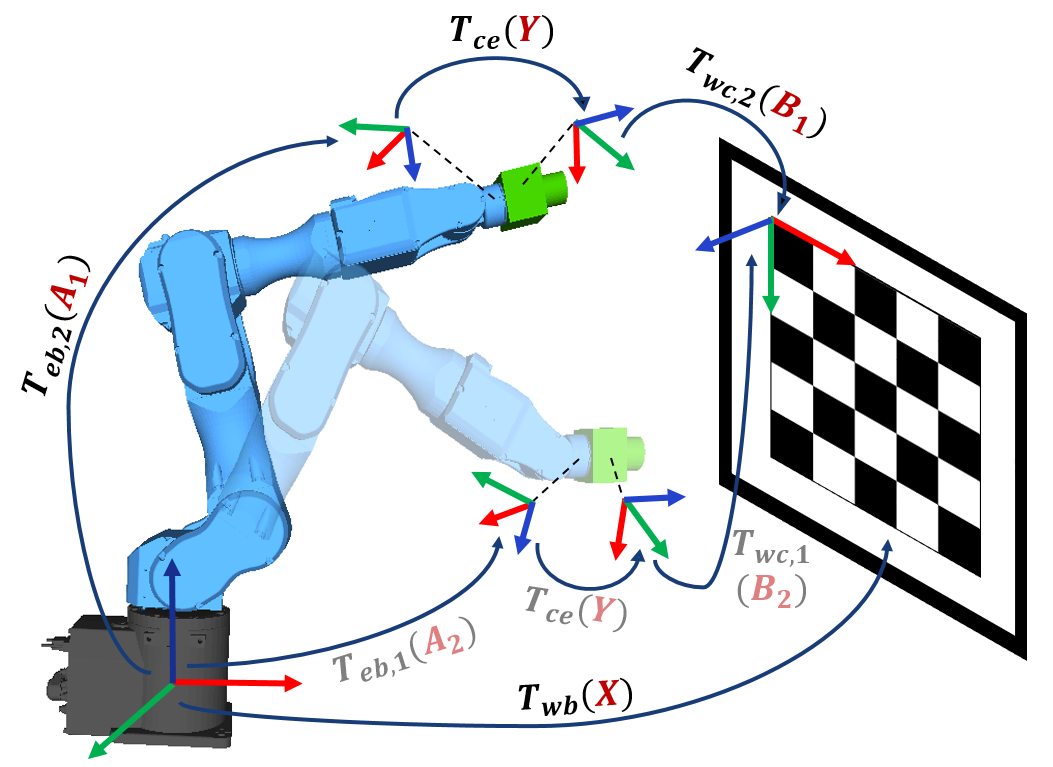}
\caption{Illustration of the robot-world-hand–eye calibration, which is formulated with the $\mathbf{AX=YB}$ mathematical representation. Multiple robot motions and camera poses are needed to ensure high calibration accuracy.}
\vspace{-0.5\baselineskip}
\label{fig_calib}
\end{figure}

As studied in~\cite{tabb2017solving,ali2019methods}, optimization-based approaches with $\mathbf{AX=YB}$ formulation achieve the highest accuracy since they use absolute measurements and are more tolerant to the measurement noise and errors. These approaches generally take the initial estimates from any fast closed-form methods, such as Tsai~\cite{tsai1989new} or Zhuang~\cite{zhuang1994simultaneous}, and iteratively minimize the formulated cost function (e.g., 2-D reprojection errors) with non-linear optimizers. Although the optimization-based solution can provide accurate calibration, it requires sufficient measurement sets to ensure high accuracy. As illustrated in Figure~\ref{fig_calib}, different robot poses are needed to create measurement sets for the optimization-based methods, and the calibration accuracy heavily relies on the selection of the measurement sets. Manual selection of the robot poses has two major limitations. First, it requires a large volume of the robot motions to collect sufficient measurements for accurate calibration. Second, even if a large volume of measurement sets is collected, it does not guarantee that a good calibration can be estimated. The addition of poor measurements may even degrade the calibration accuracy due to the biased distribution of the robot poses and measurement noise.

To this end, we propose an active vision method for eye-in-hand calibration. Our method follows the $\mathbf{AX=YB}$ formulation and uses an information theoretic next-best-view (NBV) policy~\cite{chen2011active}. For each iteration, we determine the next best robot pose from where to collect measurements, such that it brings about the maximum reduction in calibration parameter uncertainty. We evaluate our approach on a simulated dataset and verify it on a real 6-axis robot manipulator. The results demonstrate that our next-best-view approach can achieve high calibration accuracy using significantly fewer viewpoints when compared with heuristic-based baselines, such as random and maximum distance sampling strategies. The contributions are summarized as follows:
\begin{itemize}
    \item An information-theoretic formulation for the estimation of the eye-in-hand calibration parameters and prediction of the information gain for a future viewpoint.
    \item An active vision system that exploits the proposed information-theoretic formulation for rapid and accurate eye-in-hand calibration.
\end{itemize}

The rest of the paper is structured as follows. Section \ref{sec2} reviews the relevant literature. Section \ref{sec3} formulates our eye-in-hand calibration problem. Section \ref{sec4} describes the autonomous active calibration system. Section \ref{sec5} presents the evaluation results, and section \ref{sec6} concludes the paper.

\section{Related Work}
\label{sec2}
\subsection{Robotic Hand-Eye Calibration}
In robotics, hand-eye calibration is the problem of determining the rigid transformation between a camera and a robot reference frame and includes two different camera setups: eye-in-hand and eye-to-hand. The eye-in-hand calibration is the process of estimating the relative 6D pose of a robot-mounted camera with respect to the robot’s end-effector. In comparison, for eye-to-hand calibration, the camera is mounted statically, and the calibration determines the 6D pose of the camera with respect to the robot’s base. Both camera setups employ the $\mathbf{AX=XB}$ hand–eye formulation~\cite{tsai1989new,shiu1987calibration,horaud1995hand,fassi2005hand,liang2008hand} or $\mathbf{AX=YB}$ robot-world-hand–eye formulation~\cite{zhuang1994simultaneous,dornaika1998simultaneous,aiguo2010simultaneous,remy1997hand,tabb2017solving}. For both the $\mathbf{AX=XB}$ or $\mathbf{AX=YB}$ formulation, the earliest approaches solved the problem in the closed-form manner by estimating the rotation and translation separately~\cite{tsai1989new,shiu1987calibration,zhuang1994simultaneous,liang2008hand,shah2013solving}, which is also known as the separable closed-form method. Another class of closed-form methods solves the rotation and translation simultaneously~\cite{chen1991screw,dornaika1998simultaneous}. To further improve the calibration accuracy for high-precision tasks, such as robot pick-and-assembly, some recent works proposed optimization-based approaches~\cite{remy1997hand,tabb2017solving,ali2019methods}. These approaches take the initial estimates from a fast closed-form method~\cite{tsai1989new,zhuang1994simultaneous,dornaika1998simultaneous}, and iteratively minimize a cost function to yield better calibration accuracy. While the core problem has been well addressed, the majority of existing works perform the calibration in a passive way, where the robot motions and camera poses are given in advance, which limits the resulting calibration accuracy. Only a few approaches exist~\cite{rebello2017autonomous,zhang2022active} that can perform hand-eye calibration in an active manner.

\subsection{Active Vision and Next-Best-View}
Active vision~\cite{chen2011active,aloimonos1988active,bajcsy2018revisiting}, and more specifically
Next-Best-View (NBV)~\cite{connolly1985determination} refers to the approach of actively manipulating the camera to gather more informative measurements in a greedy fashion. Active vision technologies have been successfully applied to a wide range of robotic applications, such as localization~\cite{costante2018exploiting,zhang2019beyond}, 3D reconstruction~\cite{isler2016information,wu2019plant,yang2022next}, robot grasping~\cite{wu2021object,morrison2019multi} and calibration~\cite{rebello2017autonomous,zhang2022active}. Among these works, a typical scheme for NBV is to maximize the Fisher information or minimize the entropy for the robot state parameters~\cite{costante2018exploiting,zhang2019beyond,rebello2017autonomous}. Specifically, for a robot state estimation problem, the "informativeness" of a viewpoint (i.e., how much parameter uncertainties can be reduced) is quantified by the Fisher information. In~\cite{costante2018exploiting,zhang2019beyond}, the authors leverage the Fisher information to select informative trajectories to improve localization quality and avoid pose tracking loss. From these examples, the closest one to our work is from~\cite{rebello2017autonomous}, where the Fisher information is used to select the next best robot mechanism inputs for calibrating a dynamic camera cluster with one static camera~\cite{das2016calibration}. In this work, we formulate the active robot eye-in-hand calibration problem with the Fisher information objective and automate the calibration process using a next-best-view strategy.

\section{Eye-in-Hand Calibration Formulation}
\label{sec3}
In this section, we present our problem formulation for eye-in-hand calibration. We use the robot-world-hand–eye calibration formulation, $\mathbf{AX=YB}$, and solve the problem using an iterative non-linear optimization-based approach. We define the rigid transformation $\mathbf{T}$ in the $\mathbb{SE}(3)$ Lie Group:
\begin{equation}
\label{lie_group}
    \mathbb{SE}(3) := \Biggl\{\mathbf{T}=\begin{bmatrix}
\mathbf{R} & \mathbf{t} \\
\mathbf{0}^T & 1
\end{bmatrix} \: \Bigg | \:\:\: \mathbf{R}\in\mathbb{SO}(3),\mathbf{t}\in\mathbb{R}^3
\Biggl\}
\end{equation}
where $\mathbb{SO}(3)$ is the special orthogonal group (i.e., $\mathbf{R}\mathbf{R}^T = \mathbf{I}$, $\mathsf{det}(\mathbf{R})=1$). The associated Lie Algebra space to the $\mathbb{SE}(3)$ Lie Group is indicated as $\mathfrak{se}(3)$. The calibration process aims to determine two rigid transformations in $\mathbb{SE}(3)$: the transformation $\mathbf{T}_{ce}$ from the robot end-effector frame, $\mathcal{F}_e$, to the camera frame, $\mathcal{F}_c$, as well as the transformation $\mathbf{T}_{bw}$ from the world frame, $\mathcal{F}_w$, to the robot base frame, $\mathcal{F}_b$. Note that the world frame, $\mathcal{F}_w$, is defined by a static calibration board with a known marker size (e.g., a checkerboard). For each robot pose, $\mathbf{T}_{eb}$, the camera captures the pixel measurements, $\mathbf{u} \in \mathbb{R}^2$. We define the measurement set, $\mathbf{Z}_k$, from the $k^{th}$ robot pose as:
\begin{equation}
\label{equ_meas_set}
    \mathbf{Z}_k = \bigl\{ \mathbf{u}_k ,\:  \mathbf{T}_{eb,k}\bigl\}
\end{equation}
where the robot pose, $\mathbf{T}_{eb,k}$, represents the transformation from the robot base frame, $\mathcal{F}_b$, to the robot end-effector frame, $\mathcal{F}_e$, which is obtained from the robot's forward kinematics. The pixel measurement, $\mathbf{u}_k \in \mathbb{R}^2$, is the set of the 2D projections of the 3D marker points, $\mathbf{P}_{w} \in \mathbb{R}^3$, from the calibration board. 

Given the measurement set, $\mathbf{Z}_k$, estimating the transformations $\mathbf{T}_{ce}$ and $\mathbf{T}_{bw}$ can be formulated as a non-linear optimization problem. We name these two unknown transformations as the set of calibration parameters and represent them as:
\begin{equation}
\label{equ_calib_set}
    \boldsymbol{\Theta} = \bigl\{\mathbf{T}_{ce} \:\:,\:\: \mathbf{T}_{bw} \bigl\}
\end{equation}
The optimization is constructed by defining the re-projection error between the measured pixel positions, $\mathbf{u}_k$, and the projected target locations, $\mathbf{u}_k^{\prime}$, through the calibration parameters on the image plane:
\begin{align}
\label{equ_residual}
    \boldsymbol{r}^j\left(\boldsymbol{\Theta}, \mathbf{Z}_k \right) &= \boldsymbol{u}_k^j - \boldsymbol{u}_k^{j\prime}\\
    &= \boldsymbol{u}_k^j - \pi\left(\boldsymbol{T}_{cw, k}\boldsymbol{P}_w^j\right)\\
    &= \boldsymbol{u}_k^j - \pi\left(\boldsymbol{T}_{ce}\boldsymbol{T}_{eb, k}\boldsymbol{T}_{bw}\boldsymbol{P}_w^j\right)
\end{align}
where $\boldsymbol{u}_k^j$ is the measurement of the $j^{th}$ 3D marker point $\boldsymbol{P}_w^j$ from the $k^{th}$ robot pose, and $\boldsymbol{r}^j\left(\boldsymbol{\Theta}, \mathbf{Z}_k \right)$ is the residual between them. $\pi$ represents the perspective projection function for a camera model. The loss function is then defined as:
\begin{align}
\label{equ_loss}
    L\left(\boldsymbol{\Theta}\right) &= \sum_k \sum_{j=1}^{\left| \boldsymbol{P}_w \right|} \boldsymbol{r}^j\left(\boldsymbol{\Theta}, \mathbf{Z}_k \right)^T \boldsymbol{\left({\Sigma_{k}^j}\right)}^{-1}  \: \boldsymbol{r}^j\left(\boldsymbol{\Theta}, \mathbf{Z}_k \right)
\end{align}
where $\boldsymbol{\Sigma_{k}^j}$ is the corresponding measurement covariance matrix. For a calibration process, the calibration target is usually provided with high contrast, and the 2D measurements (e.g., corner detection) of the 3D marker points can be obtained accurately. Therefore, we assume the measurement noise is constant for different marker positions across all the camera viewpoints and change the loss function in Equation~(\ref{equ_loss}) to:
\begin{align}
\label{equ_loss_approx}
    L\left(\boldsymbol{\Theta}\right) &\approx \sum_k \sum_{j=1}^{\left| \boldsymbol{P}_w \right|} \boldsymbol{r}^j\left(\boldsymbol{\Theta}, \mathbf{Z}_k \right)^T \: \boldsymbol{r}^j\left(\boldsymbol{\Theta}, \mathbf{Z}_k \right)
\end{align}

To compute the optimal calibration parameters $\boldsymbol{\Theta}^{*}$, we perform an unconstrained optimization for Equation~(\ref{equ_loss_approx}) and minimize the re-projection errors over all the collected measurement sets, $\mathbf{Z}_{1:K}$. To ensure a stable solution, we can provide a good initial estimate $\overline{\boldsymbol{\Theta}}$ from any closed-form methods~\cite{zhuang1994simultaneous,aiguo2010simultaneous,shah2013solving} for the non-linear optimization. The iterative algorithms, such as Levenberg–Marquardt or Gauss-Newton, can finally be used to optimize the calibration parameters. For example, the update equation with the Gauss-Newton method is:
\begin{align}
\label{equ_optimize}
    \left( \boldsymbol{J}_{\boldsymbol{\Theta}, \boldsymbol{Z}_{1:K}}^T \boldsymbol{J}_{\boldsymbol{\Theta}, \boldsymbol{Z}_{1:K}} \right) \delta \boldsymbol{\Theta} = \boldsymbol{J}_{\boldsymbol{\Theta}, \boldsymbol{Z}_{1:K}}^T \:\: \boldsymbol{r}\left(\boldsymbol{\Theta}, \boldsymbol{Z}_{1:K}\right)
\end{align}
where $\boldsymbol{r}\left(\boldsymbol{\Theta}, \boldsymbol{Z}_{1:K}\right)$ is the residual vector over all the collected measurements $\boldsymbol{Z}_{1:K}$. The stacked Jacobian matrix $\boldsymbol{J}_{\boldsymbol{\Theta}, \boldsymbol{Z}_{1:K}}$ of the measurements is represented as:
\begin{align}
\label{equ_jacobian}
    \boldsymbol{J}_{\boldsymbol{\Theta}, \boldsymbol{Z}_{1:K}} =
    \begin{bmatrix}
    \boldsymbol{J}_{\boldsymbol{\Theta}, \boldsymbol{Z}_{1}} \\
    \vdots \\
    \boldsymbol{J}_{\boldsymbol{\Theta}, \boldsymbol{Z}_{K}}
    \end{bmatrix}
\end{align}

Each row-block of Equation~(\ref{equ_jacobian}), $\boldsymbol{J}_{\boldsymbol{\Theta}, \boldsymbol{Z}_{k}}$, corresponds to the Jacobian matrix with the $k^{th}$ measurement set. It composes of two parts:
\begin{align}
\label{equ_jacobian_two_parts}
    \boldsymbol{J}_{\boldsymbol{\Theta}, \boldsymbol{Z}_k} = \left[\boldsymbol{F}_{k} \:\:\: \boldsymbol{E}_{k} \right]
\end{align}
where Jacobian matrices $\boldsymbol{F}_{k}$ and $\boldsymbol{E}_{k}$ are the derivative of the overall cost function with respect to the poses $\boldsymbol{\xi}_{ce}$ and $\boldsymbol{\xi}_{bw}$ on the tangent space:
\begin{align}
\label{equ_jacobian_F}
    \boldsymbol{F}_{k} = -\frac{\partial \boldsymbol{u}_k^{\prime}}{\partial \boldsymbol{\xi}_{ce}} = -\frac{\partial \boldsymbol{u}_k^{\prime}}{\partial \boldsymbol{P}_{c,k}}   \frac{\partial \boldsymbol{P}_{c,k}}{\partial \boldsymbol{\xi}_{ce}}
\end{align}
\begin{align}
\label{equ_jacobian_E}
    \boldsymbol{E}_{k} = -\frac{\partial \boldsymbol{u}_k^{\prime}}{\partial \boldsymbol{\xi}_{bw}} = -\frac{\partial \boldsymbol{u}_k^{\prime}}{\partial \boldsymbol{P}_{c,k}}   \frac{\partial \boldsymbol{P}_{c,k}}{\partial \boldsymbol{P}_{e,k}}   \frac{\partial \boldsymbol{P}_{e,k}}{\partial \boldsymbol{P}_{b,k}} \frac{\partial \boldsymbol{P}_{b,k}}{\partial \boldsymbol{\xi}_{bw}}
\end{align}
where $\boldsymbol{\xi}_{ce}\:,\: \boldsymbol{\xi}_{bw} \in \mathfrak{se}(3)$ are the Lie algebra representations of the transformations $\mathbf{T}_{ce}$ and $\mathbf{T}_{bw}$, respectively. The Hessian, $\boldsymbol{J}_{\boldsymbol{\Theta}, \boldsymbol{Z}_k}^T \boldsymbol{J}_{\boldsymbol{\Theta}, \boldsymbol{Z}_k}$ is constructed by:
\begin{align}
\label{equ_hessian}
\boldsymbol{J}_{\boldsymbol{\Theta}, \boldsymbol{Z}_k}^T \boldsymbol{J}_{\boldsymbol{\Theta}, \boldsymbol{Z}_k} = \begin{bmatrix}
\boldsymbol{F}_{k}^T \boldsymbol{F}_{k} & \boldsymbol{F}_{k}^T \boldsymbol{E}_{k}\\
\boldsymbol{E}_{k}^T \boldsymbol{F}_{k} & \boldsymbol{E}_{k}^T \boldsymbol{E}_{k}
\end{bmatrix}
\end{align}
Note that, for the multiple measurement sets, $\boldsymbol{Z}_{1:K}$, the Hessian, $\boldsymbol{J}_{\boldsymbol{\Theta}, \boldsymbol{Z}_{1:K}}^T \boldsymbol{J}_{\boldsymbol{\Theta}, \boldsymbol{Z}_{1:K}}$, is a $12 \times 12$, square matrix. The operation complexity of its inversion is constant.

\section{Active Calibration using Next-Best-View}
\label{sec4}
\begin{figure}[t]
\centering
  \includegraphics[width=\linewidth]{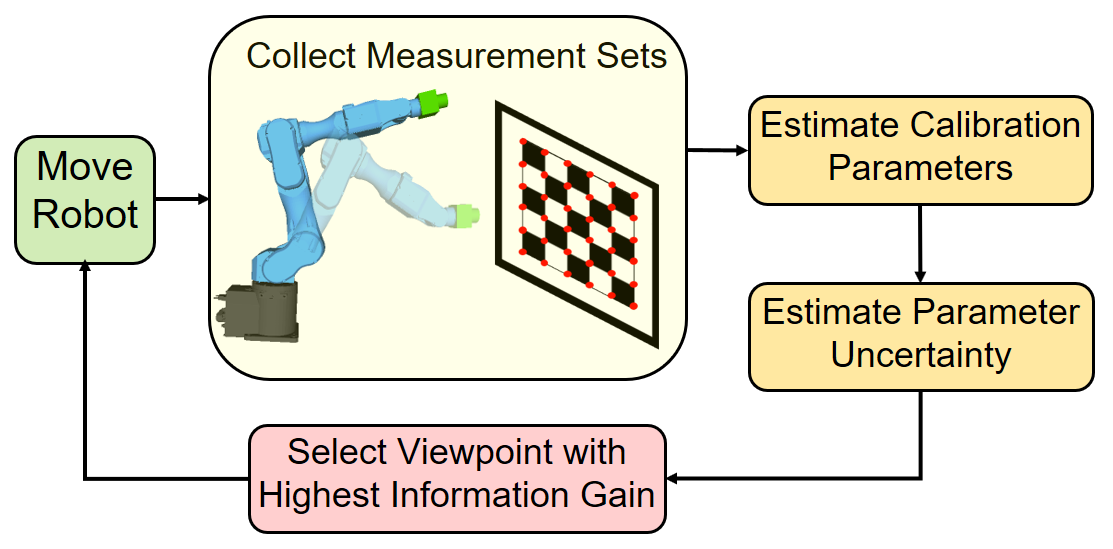}
\caption{Method overview of active eye-in-hand calibration. Given the collected measurement sets, our system first estimates the calibration parameters and the corresponding uncertainty. The next robot pose is selected so that the predicted information gain is maximized, which improves the accuracy of the calibration parameters.}
\label{fig_pipeline}
\end{figure}

In Section~\ref{sec3}, we formulate the eye-in-hand calibration as a non-linear optimization problem and solve it using the iterative approaches. However, the calibration accuracy relies heavily on the collected measurement sets from the selected robot poses. Even if a large volume of the measurement sets is collected, it does not ensure calibration accuracy due to the biased distribution of the robot poses, redundant measurements, and possible measurement noises. Hence, in this section, we present our active calibration process that can estimate the uncertainty of the calibration parameters and predict the next-best-view for maximizing the information gain (uncertainty reduction). An overview of our proposed active calibration system is illustrated in Figure~\ref{fig_pipeline}.

\subsection{Initialization and Uncertainty Estimation}
We initialize our active calibration with a collection of measurement sets $\mathbf{Z}_{1:K}$ from $K$ viewpoints. To uniquely determine the set of the calibration parameters $\boldsymbol{\Theta}$ and provide the initial estimate, at least $K=3$ sets with non-parallel rotation axes are required~\cite{strobl2006optimal}. We then perform the iterative optimization, described in Section~\ref{sec3}, to refine the calibration parameters. To assess the calibration quality with collected measurement sets, we assume the set of the calibration parameter is distributed with a uni-modal Gaussian and compute its covariance matrix, $\Sigma_{\boldsymbol{\Theta}, \boldsymbol{Z}_{1:K}}$, with a first-order approximation of the Fisher information matrix (FIM):
\begin{align}
\label{equ_FIM}
\Sigma_{\boldsymbol{\Theta}, \boldsymbol{Z}_{1:K}} &= {\left( \boldsymbol{J}_{\boldsymbol{\Theta}, \boldsymbol{Z}_{1:K}}^T \: \boldsymbol{J}_{\boldsymbol{\Theta}, \boldsymbol{Z}_{1:K}} \right)}^{-1}
\end{align}

Such an inverse of the FIM defines the Cramer-Rao lower bound, which is the smallest covariance that can be achieved by an unbiased estimator. To quantify the calibration parameter uncertainty, we use the differential entropy $h_e\left( \Sigma_{\boldsymbol{\Theta}, \boldsymbol{Z}_{1:K}} \right)$:
\begin{align}
\label{equ_differential_entropy}
h_e\left( \Sigma_{\boldsymbol{\Theta}, \boldsymbol{Z}_{1:K}} \right) = \frac{1}{2} \ln{\left(\left(2\pi e\right)^n \left| \Sigma_{\boldsymbol{\Theta}, \boldsymbol{Z}_{1:K}} \right| \right)}
\end{align}
Note that the uncertainty computation is not limited to the differential entropy and can be superseded by other metrics, such as the trace~\cite{costante2018exploiting} or the sum of the eigenvalues~\cite{kiciroglu2020activemocap} of the covariance matrix.

\subsection{Next-Best-View Prediction}
To improve the quality of the calibration parameters, we aim to find the next best robot forward kinematics $\boldsymbol{\theta}^{*}$ from a set of candidates, $\{\boldsymbol{\theta}\}$, that governs the robot pose $\mathbf{T}_{eb}$ and will be used to maximize the information gain of the calibration parameters. The information gain is defined as the uncertainty (entropy) reduction after including the measurement set from a candidate robot kinematics. For a candidate robot kinematics, $\boldsymbol{\theta}$, after including the measurement set, $\overline{\mathbf{Z}}$, the stacked Jacobian matrix becomes:
\begin{align}
\label{equ_jacobian_NBV}
    \boldsymbol{J}_{\Theta, \widehat{\boldsymbol{Z}}} =
    \begin{bmatrix}
    \boldsymbol{J}_{\Theta, \boldsymbol{Z}_{1:K}} \\
    \boldsymbol{J}_{\Theta, \overline{\boldsymbol{Z}}}
    \end{bmatrix}    
\end{align}

\begin{figure}[t]
\centering
  \includegraphics[width=\linewidth]{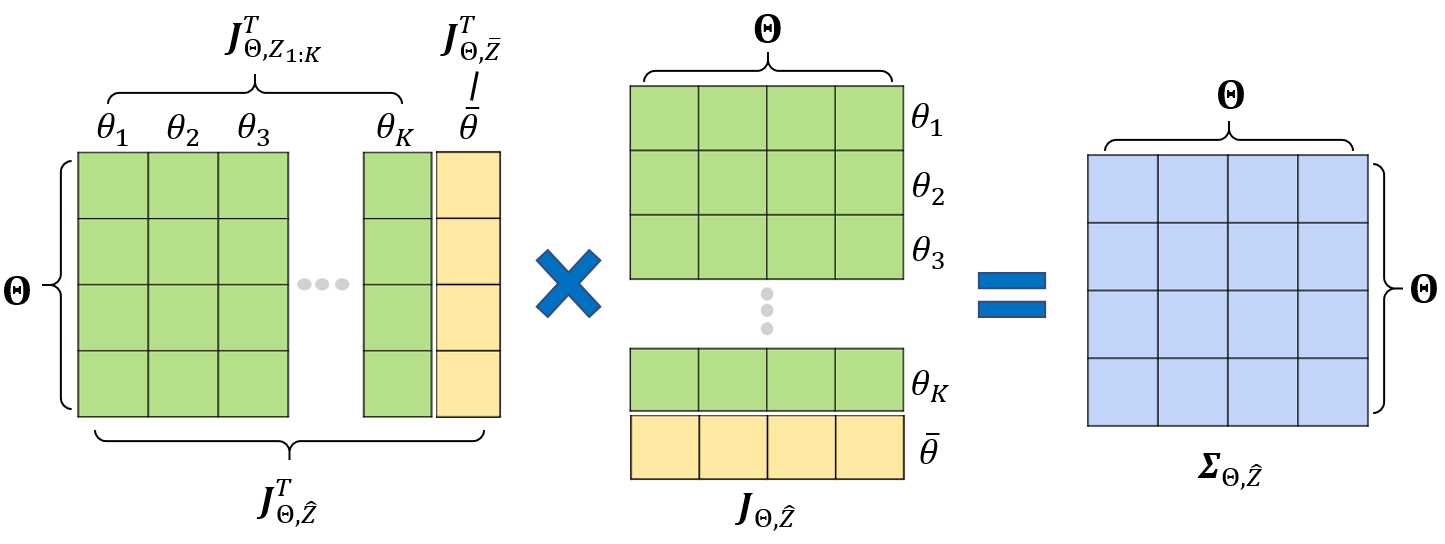}
\caption{Illustration of the prediction of the covariance matrix (blue square matrix), $\Sigma_{\boldsymbol{\Theta}, \widehat{\boldsymbol{Z}}}$, if adding the measurement from a candidate robot kinematics $\overline{\boldsymbol{\theta}}$. The predicted Jacobian matrix, $\boldsymbol{J}_{\Theta, \widehat{\boldsymbol{Z}}}$, is composed of two parts: the stacked Jacobian matrix (green block), $\boldsymbol{J}_{\Theta, \boldsymbol{Z}_{1:K}}$, from the collected measurement sets, $\boldsymbol{Z}_{1:K}$, as well as the computed Jacobian (yellow vector), $\boldsymbol{J}_{\Theta, \overline{\boldsymbol{Z}}}$, for a candidate robot kinematics $\overline{\boldsymbol{\theta}}$ before actually moving the robot.}
\label{fig_nbv}
\end{figure}

where $\widehat{\boldsymbol{Z}} = \bigl\{\boldsymbol{Z}_{1:K}, \overline{\boldsymbol{Z}}\bigl\}$ represents the set of measurements from the robot forward kinematics $\boldsymbol{\theta}_{1:K}$ and the candidate kinematics $\overline{\boldsymbol{\theta}}$. Using the FIM approximation, the parameter covariance can be calculated as follows:
\begin{align}
\label{equ_FIM_nbv}
\Sigma_{\boldsymbol{\Theta}, \widehat{\boldsymbol{Z}}} = {\left( \boldsymbol{J}_{\boldsymbol{\Theta}, \widehat{\boldsymbol{Z}}}^T \:\: \boldsymbol{J}_{\boldsymbol{\Theta}, \widehat{\boldsymbol{Z}}} \right)}^{-1}
\end{align}
Note that we compute the Jacobian matrix (Equation~(\ref{equ_jacobian_NBV})) and the predicted parameter covariance (Equation~(\ref{equ_FIM_nbv})) before actually applying the robot kinematics $\overline{\boldsymbol{\theta}}$ and including the measurement $\overline{\mathbf{Z}}$. The computation of the Equation~(\ref{equ_jacobian_NBV}) is based on the estimated calibration parameters using measurement sets $\boldsymbol{Z}_{1:K}$ only.

\begin{algorithm}[t]
   \SetKwInOut{Input}{Input}
  \SetKwInOut{Output}{Output}
  \DontPrintSemicolon
  % Option
  \Input{Initial measurement sets $\boldsymbol{Z}_{1:K}$; a set of robot candidate kinematics $\{\boldsymbol{\theta}\}$.}
  \Output{Optimized calibration parameter $\boldsymbol{\Theta}^{*}$.}
 Initial estimation of the calibration parameter $\boldsymbol{\Theta}$ using the closed-form solution from the collection of the measurement sets $\boldsymbol{Z}_{1:K}$;\; 
 Optimize the calibration parameter $\boldsymbol{\Theta}^{*}$ from the collection of the measurement sets, $\boldsymbol{Z}_{1:K}$ (Sec~\ref{sec3});\; 
 Compute the parameter covariance, $\boldsymbol{\Sigma}_{\boldsymbol{\Theta}, \boldsymbol{Z}_{1:K}}$, using equation~(\ref{equ_FIM});\; 
 Compute the information gain ${\boldsymbol{I}}_{\boldsymbol{\Theta}, \overline{\boldsymbol{Z}}}$ for each $\overline{\boldsymbol{\theta}}$ in $\{\boldsymbol{\theta}\}$ using equation~(\ref{equ_IG});\; Determine the NBV, $\boldsymbol{\theta}^{*}$, using equation~(\ref{equ_IG_max}) and apply the kinematics $\boldsymbol{\theta}^{*}$ to the robot;\;
 Collect the new measurement set $\mathbf{Z}^{*}$ and update the total collection using equation~(\ref{equ_NBV});\;
 Check if the termination condition is fulfilled;\;
 If the condition is not fulfilled: Repeat steps 2-7.
 \caption{Next-Best-View for Active Robot Eye-in-Hand Calibration}
 \label{alg1}
\end{algorithm}

For a robot kinematics $\overline{\boldsymbol{\theta}}$, we define the corresponding information gain ${I}_{\boldsymbol{\Theta}, \overline{\boldsymbol{Z}}}$ as the entropy reduction of the parameter covariance:
\begin{align}
\label{equ_IG}
{I}_{\boldsymbol{\Theta}, \overline{\boldsymbol{Z}}} = h_e\left(\Sigma_{\boldsymbol{\Theta}, \boldsymbol{Z}_{1:K}}\right) - h_e\left(\Sigma_{\boldsymbol{\Theta}, \widehat{\boldsymbol{Z}}}\right)
\end{align}
To find the next-best-view, $\boldsymbol{\theta}^{*}$, we maximize the information gain over the entire candidate set, $\{\boldsymbol{\theta}\}$:
\begin{align}
\label{equ_IG_max}
\boldsymbol{\theta}^{*} = \argmax_{\boldsymbol{\theta}} {I}_{\boldsymbol{\Theta}, \overline{\boldsymbol{Z}}}
\end{align}

Once the next-best-view, $\boldsymbol{\theta}^{*}$, is determined, we apply the robot motion. A new measurement set $\mathbf{Z}^{*}$ is collected and the total collection $\mathbf{Z}_{1:K+1}$ will be updated as:
\begin{equation}
\label{equ_NBV}
    \mathbf{Z}_{1:K} \cup \mathbf{Z}^{*} \rightarrow \mathbf{Z}_{1:K+1}
\end{equation}
We finally optimize the calibration parameters $\boldsymbol{\Theta}^{*}$ using the newly updated measurement sets, $\mathbf{Z}_{1:K+1}$, and predict the next-best-view again. We repeat this process until it reaches user-defined criteria, i.e., after a fixed number of iterations or when the highest information gain of a subsequent robot kinematics falls below a user-defined threshold:
\begin{equation}
\label{equ15}
    \boldsymbol{I}_{\boldsymbol{\Theta}, \overline{Z}} < \boldsymbol{I_{\tau}}
\end{equation}
The architecture of NBV for eye-in-hand calibration is shown in Algorithm \ref{alg1}.

\begin{figure}[t]
\centering
\begin{subfigure}{0.15\textwidth}
  \includegraphics[width=\linewidth]{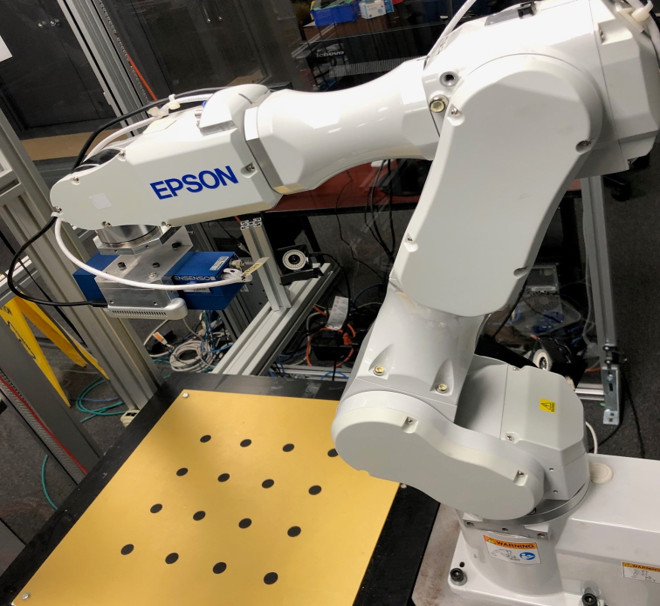}
  \caption{}
  \label{fig_robot}
\end{subfigure}
\begin{subfigure}{0.15\textwidth}
    \includegraphics[width=\linewidth]{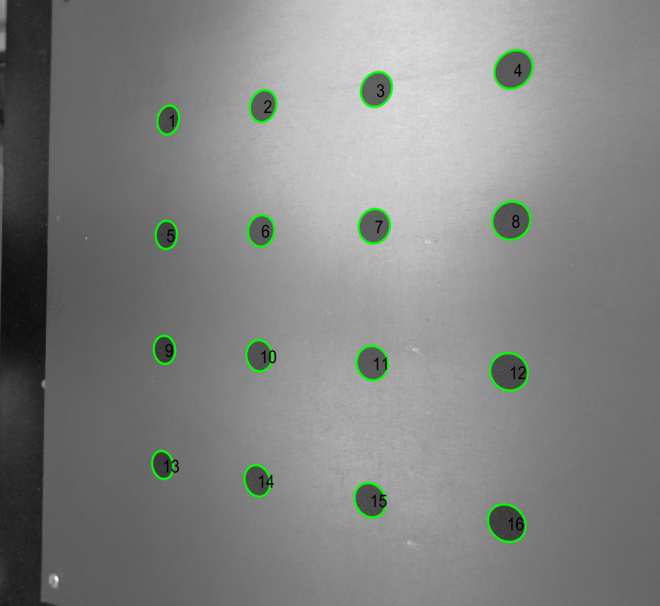}
    \caption{}
    \label{fig_pattern1}
\end{subfigure}
\begin{subfigure}{0.15\textwidth}
  \includegraphics[width=\linewidth]{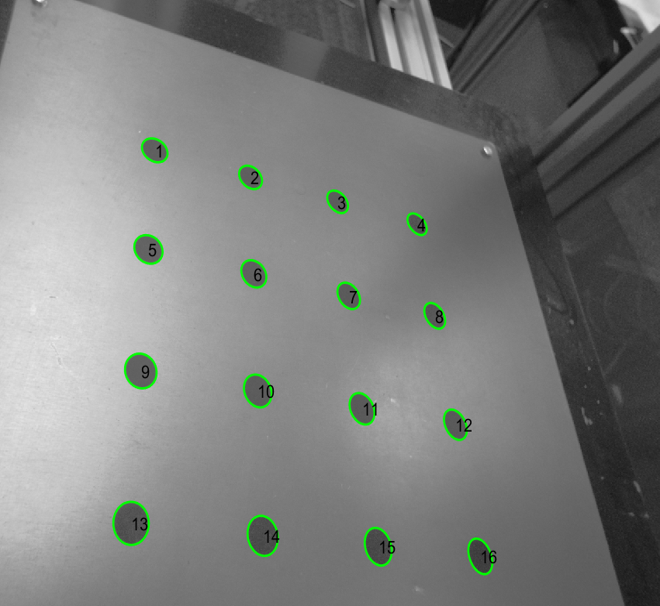}
  \caption{}
  \label{fig_pattern3}
\end{subfigure}
\begin{subfigure}{0.23\textwidth}
  \includegraphics[width=\linewidth]{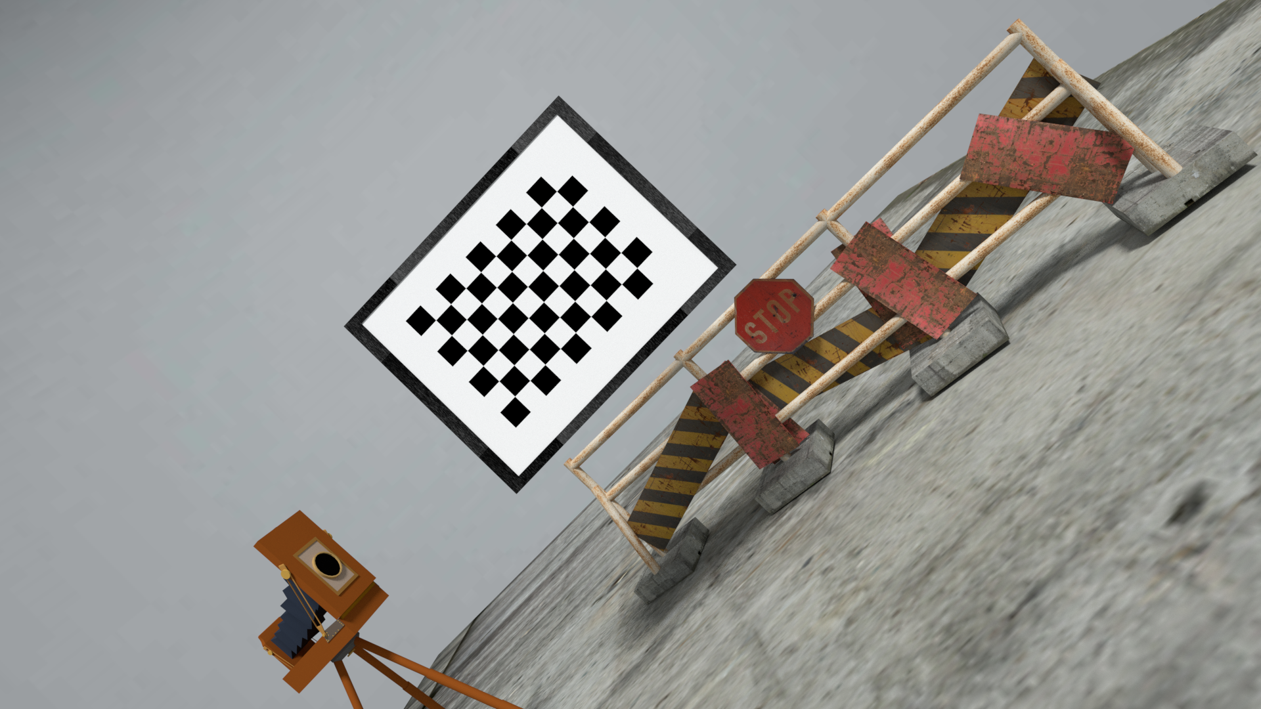}
  \caption{}
  \label{fig_sim1}
\end{subfigure}
\begin{subfigure}{0.23\textwidth}
  \includegraphics[width=\linewidth]{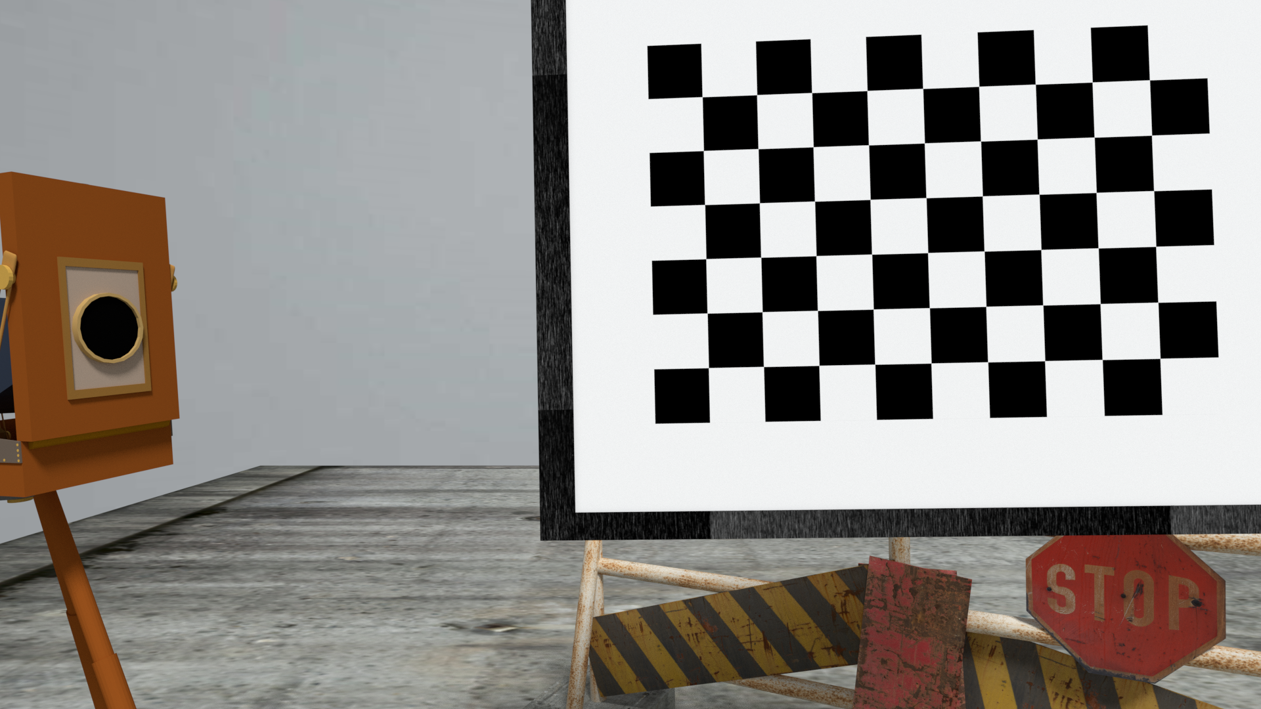}
  \caption{}
  \label{fig_sim2}
\end{subfigure}
\caption{(a) The experimental setup for acquiring the real dataset: The EPSON C4L 6-Axis robot manipulator and the calibration board, we mount an IDS Ensenso N35 camera on the robot's end-effector. (b)-(c) The examples of the captured images using the Ensenso N35 camera from different viewpoints. (d)-(e) The examples of the synthetic images from the simulated dataset~\cite{ali2019methods}.}
\label{fig_setup}
\vspace{-0.5\baselineskip}
\end{figure}

\section{Experiments}
\label{sec5}
\subsection{Datasets}
\textbf{Real-World Dataset.} To demonstrate the effectiveness of our proposed approach, we capture a real dataset using an EPSON C4L 6-Axis robot manipulator with an industrial Ensenso N35 camera. As shown in Figure~\ref{fig_robot}, we mount the camera on the end-effector of the robot arm and move the robot to capture the monochrome images with varying robot poses. In our experiments, we use a precisely manufactured calibration board with a $4\times4$ symmetrical circle grid pattern. We calibrated the camera's intrinsic parameters using the calibration toolbox provided by the camera vendor. We program the robot arm to move the camera to different viewpoints. Since the calibration pattern is symmetrical, we make the robot end-effector stay pointed towards the center of the workstation, and the camera in-plane rotation remains unchanged. The examples of the captured calibration images are shown in Figure~\ref{fig_pattern1}-\ref{fig_pattern3}. With such a workstation setup, we captured a total of seven sets. Each calibration set includes one training set and one validation set. We perform the calibration using the training set only and evaluate the calibration results on the validation set.
\vspace{0.4\baselineskip}

\textbf{Simulated Dataset.} The real dataset can represent the true uncertainties (e.g., measurement noise) for a calibration system in the real world. However, it is not possible to acquire the ground truth information for quantifying the absolute pose errors. The main advantage of using a simulated dataset is the availability of ground truth transformations (e.g., robot end-effector to camera frame, robot base to world frame). In our experiments, we use a public dataset~\cite{ali2019methods}. It includes three simulated datasets with a different number of robot poses and synthetic images. The dataset is injected with pseudo-realistic robot pose noise and visual noise for synthetic images.

\begin{figure}[t]
\centering
  \includegraphics[width=0.9\linewidth]{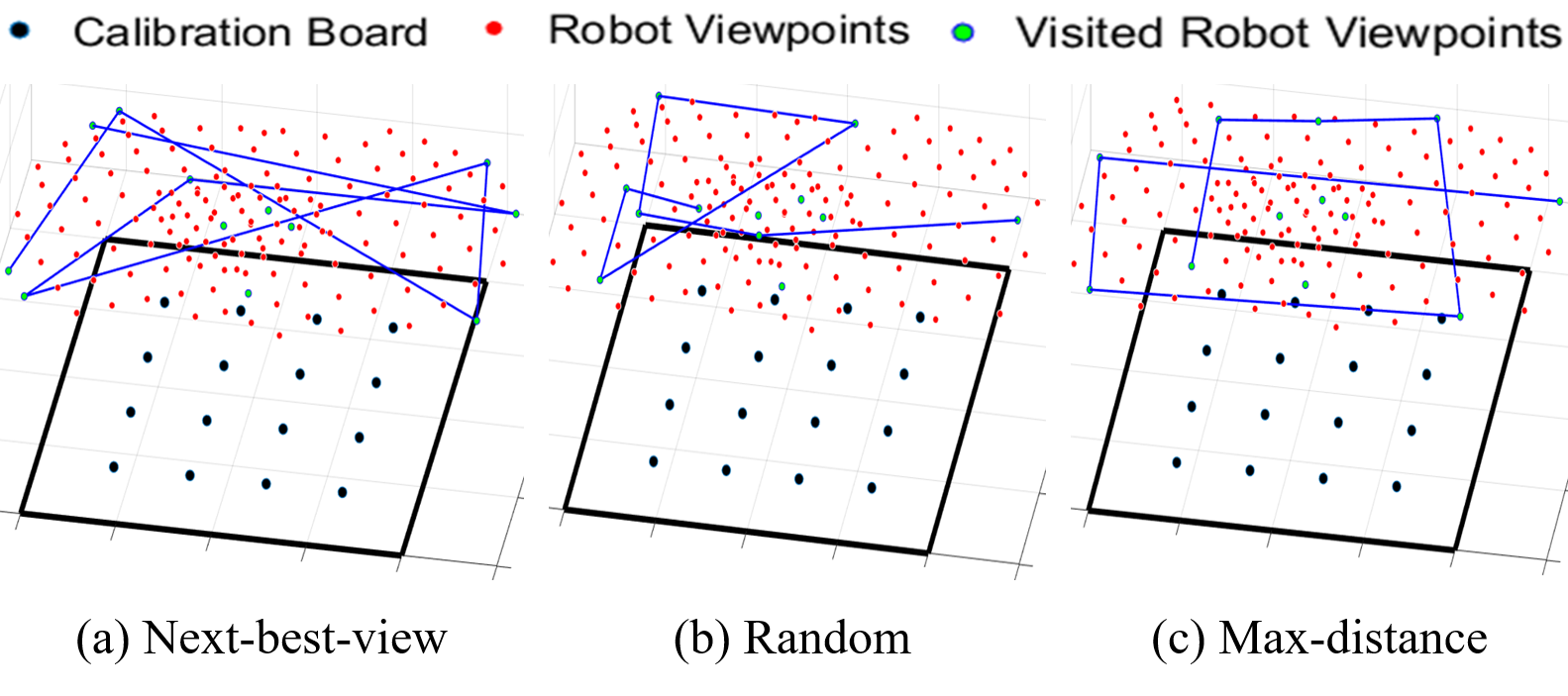}
\caption{Trajectories found by our strategy, along with random and max-distance baselines. The trajectories generated by our algorithm have large baselines between viewpoints with respect to the calibration board.}
\label{fig_baseline}
\vspace{-0.5\baselineskip}
\end{figure}

\subsection{Baselines and Evaluation Metrics}
\label{sec_baseline_metric}
For robot hand-eye calibration, existing works use heuristic-based policies, such as random selection or linear spacing, to select viewpoints. To demonstrate the effectiveness of our approach, we compare our system against two heuristic-based viewpoint selection strategies. The first baseline, "Random", selects random robot poses from the set of candidates. The second baseline, "Maximum Distance", moves the robot end-effector to the position of the furthest distance from previous positions in the Euclidean space $\mathbb{R}^3$. Figure~\ref{fig_baseline} depicts the trajectories generated by our approach and these two baselines. In our experiments, we assume the robot can teleport from one location to the next without the path planning restriction. This will allow us to explore the theoretical upper bounds of our method.

To evaluate the results on the simulated datasets, we take two metrics from~\cite{ali2019methods}. We use absolute translation error (mm), $e_\mathbf{at}$, and absolute rotation error (deg), $e_\mathbf{aR}$, to quantify the absolute pose errors:
\begin{equation}
\label{equ_metric_aT}
    \Delta \mathbf{T}_{ce} =  \mathbf{\check{T}}_{ce}^{-1} \: \mathbf{\overline{T}}_{ce}
\end{equation}
\begin{equation}
\label{equ_metric_atR}
    e_\mathbf{at} = d\left(\Delta \mathbf{T}_{ce}\right), \:\:\: e_\mathbf{aR} = \angle\left(\Delta \mathbf{T}_{ce}\right)
\end{equation}
where $\mathbf{\check{T}}_{ce}$ and $\mathbf{\overline{T}}_{ce}$ are the estimated and ground truth transformation (from robot end-effector to camera frame), respectively. The function $d\left( \Delta \mathbf{T} \right)$ extracts the translation part from $\Delta \mathbf{T}$ and computes the Euclidean norm. $\angle\left( \Delta \mathbf{T} \right)$ extracts the rotation part from $\Delta \mathbf{T}$ and gets the absolute angle value with angle-axis representation with Lie algebra representation in $\mathfrak{so}(3)$. 

For the real dataset, due to the missing ground truth poses, we use three metrics from~\cite{ali2019methods} for the evaluation: relative translation error (mm), $e_\mathbf{rt}$, relative rotation error (deg), $e_\mathbf{rR}$, and reprojection error (px), $e_\mathbf{RMSE}$. The relative translation and rotation errors are derived from $\mathbf{AX=YB}$ formulation:
\begin{equation}
\label{equ_metric_T}
    \Delta \mathbf{T}_k = \mathbf{\check{T}}_{bw} \: \mathbf{T}_{wc,k} \: \mathbf{\check{T}}_{ce} \: \mathbf{T}_{eb,k}
\end{equation}
\begin{equation}
\label{equ_metric_t}
    e_\mathbf{rt} =  \frac{1}{K} \sum_{k=1}^K d\left( \Delta \mathbf{T}_k \right), \:\:\: e_\mathbf{rR} =  \frac{1}{K} \sum_{k=1}^K \angle\left( \Delta \mathbf{T}_k \right)
\end{equation}
where $\mathbf{\check{T}}_{bw}$ and $\mathbf{\check{T}}_{ce}$ are the estimated calibration parameters. The per-frame camera pose $\mathbf{T}_{wc,k}$ is obtained by solving the perspective-n-point (PnP) problem. The metric re-projection error, $e_\mathbf{RMSE}$, is the re-projection root mean squared error between the measured pixel positions and the projected locations over the test set:
\begin{equation}
\label{equ_metric_T}
    e_\mathbf{RMSE} = \sqrt{ \frac{1}{K-1} \sum_{k=1}^K {\lVert \boldsymbol{u}_k - \pi\left(\boldsymbol{T}_{ce}^*\boldsymbol{T}_{eb, k}\boldsymbol{T}_{bw}^*\boldsymbol{P}_w\right) \rVert}_2^2 }
\end{equation}
where $\boldsymbol{u}_k$ is the pixel measurement of the 3D marker point set $\boldsymbol{P}_w$ from the robot pose $\boldsymbol{T}_{eb, k}$.

\begin{figure*}[t]
\centering
\begin{subfigure}{0.245\textwidth}
  \includegraphics[width=\linewidth]{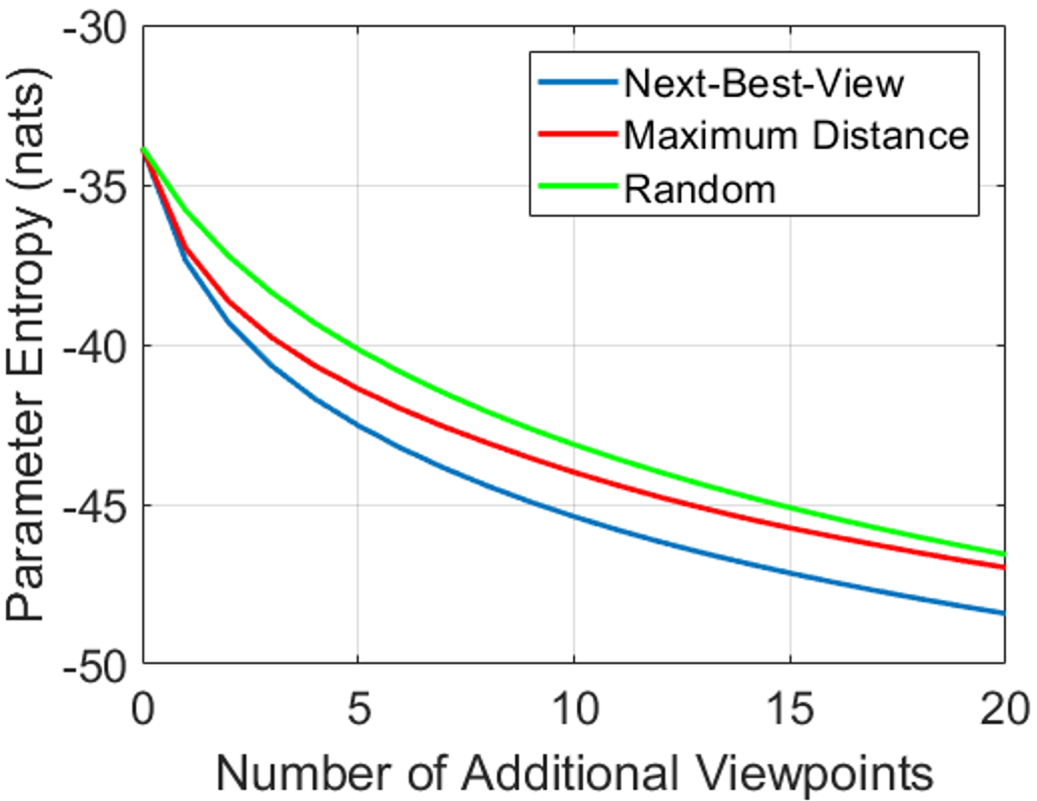}
  \caption{}
  \label{fig_cov}
\end{subfigure}
\begin{subfigure}{0.245\textwidth}
  \includegraphics[width=\linewidth]{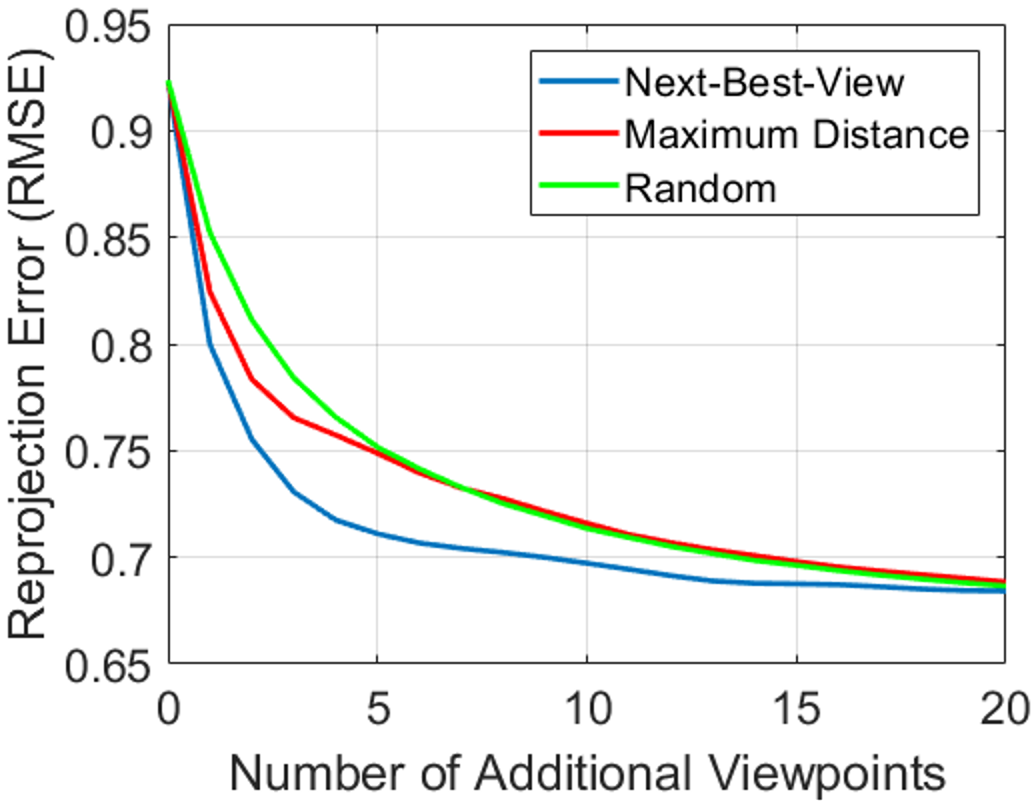}
  \caption{}
  \label{fig_cov}
\end{subfigure}
\begin{subfigure}{0.245\textwidth}
  \includegraphics[width=\linewidth]{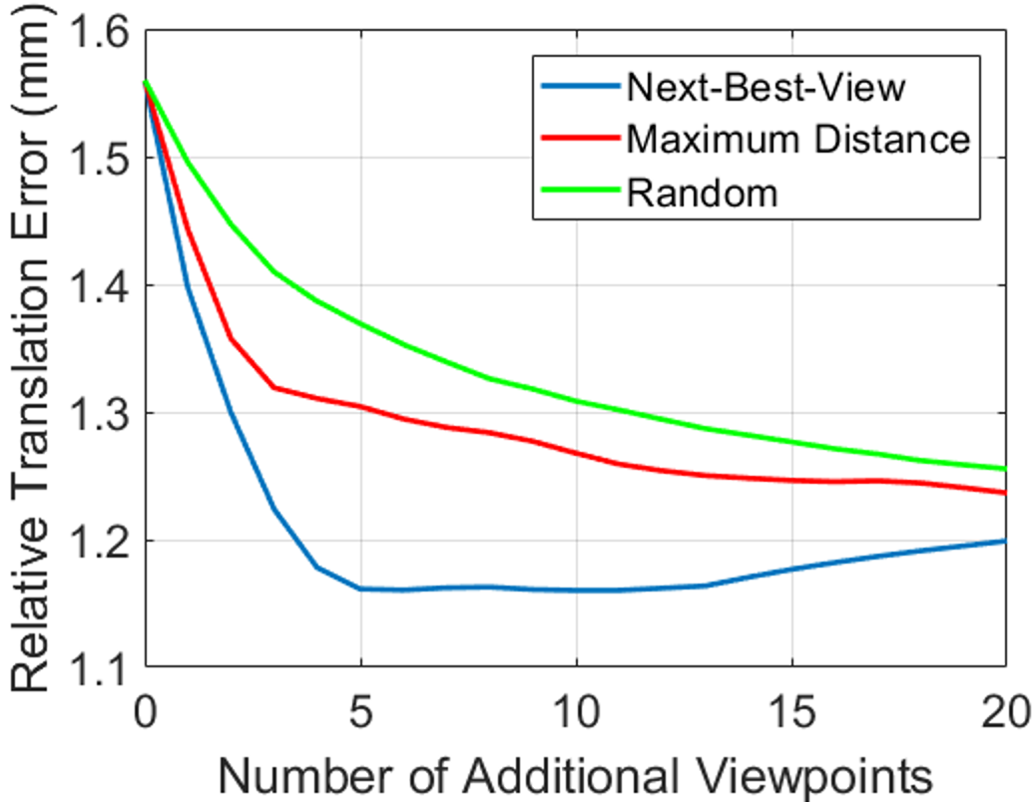}
  \caption{}
  \label{fig_cov}
\end{subfigure}
\begin{subfigure}{0.245\textwidth}
  \includegraphics[width=\linewidth]{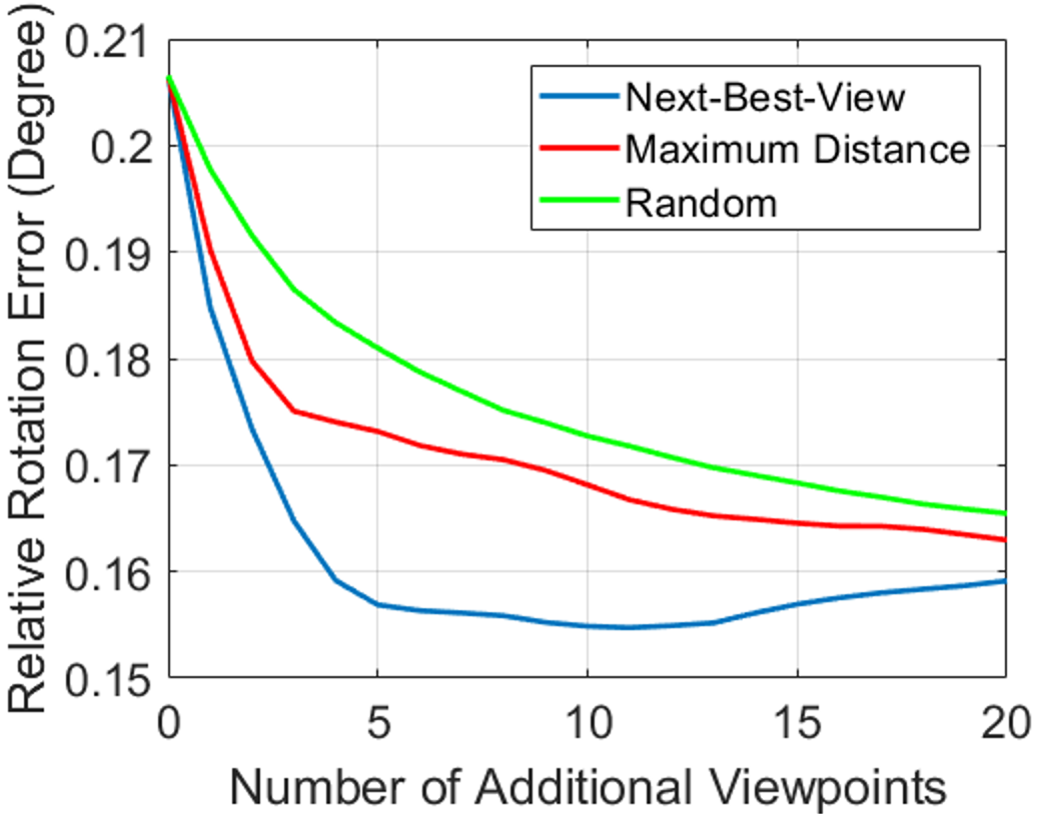}
  \caption{}
  \label{fig_cov}
\end{subfigure}
\caption{Comparison of the evaluation results versus the number of additional viewpoints for the baselines and our NBV approach. (a) Entropy. (b) Re-projection RMSE. (c) Relative Translation error. (d) Relative Rotation error. It is clear that our approach can achieve low parameter uncertainty and high test accuracy with fewer robot viewpoints.}
\vspace{-0.5\baselineskip}
\label{fig_results}
\end{figure*}

\subsection{Results}
For our system and the baselines, we initialize the calibration parameters with the same set of robot poses (3 sets in our experiments). Table~\ref{tab_result_sim} shows the experimental results on the simulated dataset~\cite{ali2019methods}. We can observe that, although the proposed approach is not advantageous to the baselines in terms of the absolute rotation error, it can achieve much higher accuracy for the translation estimation using the same number of the collected measurement sets (28.4\% and 37.7\% error reduction compared to the "random" and "max-distance" baselines, respectively).

\begin{table}[]
\resizebox{0.485\textwidth}{!}{
\begin{tabular}{|c|c|c|}
\hline
\backslashbox{Method}{Metric}        & \begin{tabular}[c]{@{}c@{}}Absolute Translation\\ Error (mm)\end{tabular} & \begin{tabular}[c]{@{}c@{}}Absolute Rotation\\ Error (deg)\end{tabular} \\ \hline
Random       & 4.12                                                                      & 0.016                                                                  \\ \hline
Max-Distance & 4.74                                                                      & \textbf{0.014}                                                                  \\ \hline
Proposed NBV & \textbf{2.95}                                                             & 0.015                                                        \\ \hline
\end{tabular}}
\caption{Evaluation results on the simulated dataset~\cite{ali2019methods}, evaluated with the two metrics (absolute translation and rotation errors), described in Section~\ref{sec_baseline_metric}. We set the maximum number of additional robot poses to 5.}
\label{tab_result_sim}
\end{table}

For the real dataset, we performed the evaluation on the collected validation sets, which were not used in the calibration process. Figure~\ref{fig_results} presents the results for the baselines and our proposed NBV system when using different metrics. It can be seen that, with the first few additional viewpoints, the "max-distance" achieves higher accuracy than the "random" policy and has a similar performance to our proposed NBV approach. However, the performance using both "max-distance" and "random" policies becomes less obvious when visiting more viewpoints. Compared to the baselines, our approach is able to achieve the same level of parameter uncertainties and accuracy with much fewer robot poses. Moreover, we measure the Pearson correlation coefficient~\cite{freedman2007statistics} between our predicted information gain and the reduction of the re-projection error. As illustrated in Figure~\ref{fig_nbv}, the predicted information gain accurately reflects the true error reduction, thus making them well suited to our goal. Table~\ref{tab_result} further demonstrates the advantage of our approach with the same number of additional robot poses. To obtain the results, we use 5 additional measurement sets for each view selection strategy. We can see that, compared to the baselines, our proposed NBV achieves the lowest error for all three metrics.

\begin{figure}[t]
\centering
  \includegraphics[width=\linewidth]{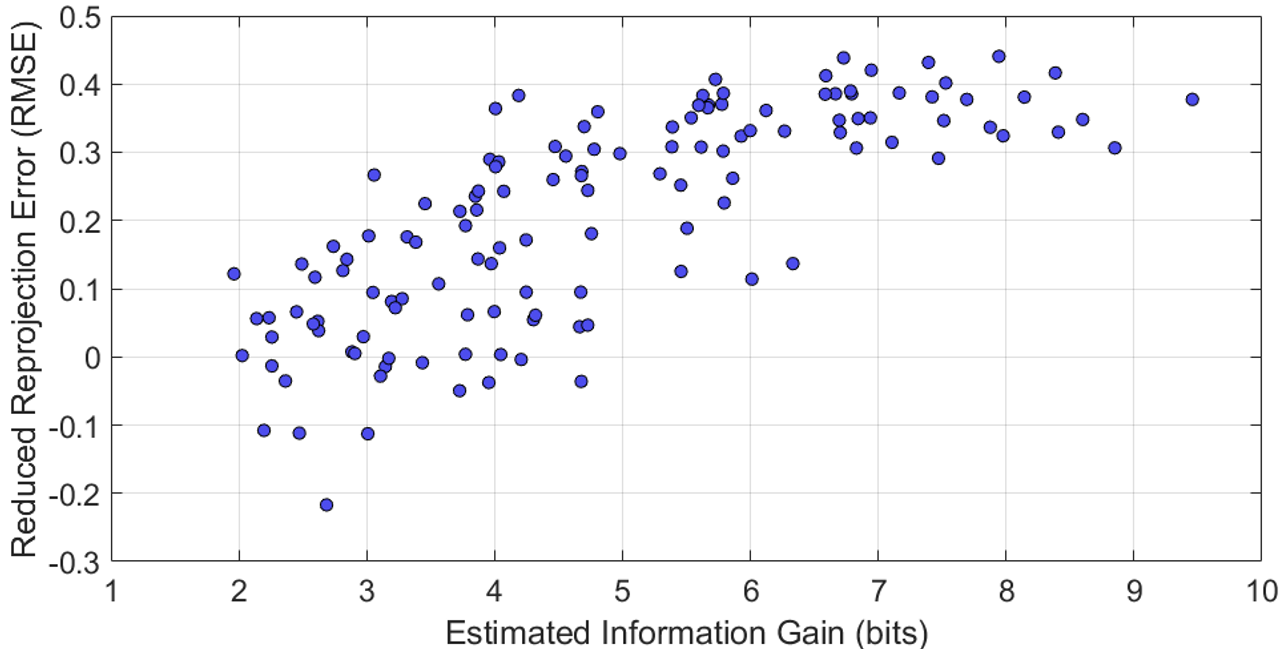}
\caption{The predicted information gain across potential robot poses compared with the reduced re-projection error (RMSE) if we were actually to apply these poses. Each blue dot represents a robot pose. We compute the Pearson correlation coefficient (\textbf{0.76} in this example). Our predicted information gain correlates well with the true reduction of the re-projection error.}
\label{fig_nbv}
\end{figure}

\section{Conclusion}
\label{sec6}
In this work, we have presented a viewpoint selection framework for eye-in-hand calibration. We first formulate the calibration problem with the $\mathbf{AX=YB}$, robot-world-hand–eye representation, and estimate the set of calibration parameters using a non-linear optimization process. Our approach estimates the parameter uncertainty and predicts the information gain of future measurement sets. This allows us to reduce the parameter uncertainty by selecting the robot kinematics with the highest information gain. We evaluate our method on a synthetic dataset and a real robot manipulator. The results demonstrate that our proposed next-best-view approach can achieve high calibration accuracy with much fewer robot poses when compared against baselines that use heuristic-based policies.

\begin{table}[]
\resizebox{0.485\textwidth}{!}{
\begin{tabular}{|c|c|c|c|}
\hline
\backslashbox{Method}{Metric}                        & \begin{tabular}[c]{@{}c@{}}Reprojection\\ Error (px)\end{tabular} & \begin{tabular}[c]{@{}c@{}}Relative\\ Translation\\  Error (mm)\end{tabular} & \begin{tabular}[c]{@{}c@{}}Relative\\ Rotation \\ Error (deg)\end{tabular} \\ \hline
Random                        & 0.766                                                              & 1.386                                                              & 0.183                                                            \\ \hline
Max-Distance                  & 0.757                                                              & 1.310                                                              & 0.174                                                            \\ \hline
Proposed NBV & {\textbf{0.717}}                                    & {\textbf{1.178}}                                    & {\textbf{0.159}}                                                                          \\ \hline
\end{tabular}}
\caption{Evaluation results on the real dataset, evaluated with the three metrics (reprojection error, relative translation, and rotation errors). The maximum number of additional robot poses is set to 5.}
\label{tab_result}
\end{table}


\section*{Acknowledgment}
This work was supported by Epson Canada Ltd.



%\bibliographystyle{IEEEtran}

\bibliographystyle{ieeetr}
\bibliography{bibliography.bib}

\end{document}